\documentclass[11pt]{article}
\usepackage{coling2018}
\usepackage{times}
\usepackage{url}
\usepackage{latexsym}

\usepackage{graphicx}
\usepackage{wrapfig}
\usepackage{url}
\usepackage{amsmath}
\usepackage{amssymb}
\usepackage{color,xcolor,colortbl}
\usepackage{enumitem}
\usepackage{algorithm}
\usepackage{algorithmic}
\usepackage[font={small}]{caption}
\usepackage{bm}
\usepackage{booktabs}
\usepackage{multirow}

\DeclareMathOperator*{\argmax}{arg\,max\,}
\newcommand\numberthis{\addtocounter{equation}{1}\tag{\theequation}}

\definecolor{darkblue}{rgb}{0.0, 0.0, 0.55}

\title{Abstract Meaning Representation for Multi-Document Summarization}

\author{Kexin Liao \quad Logan Lebanoff \quad Fei Liu\\ 
  Computer Science Department\\
  University of Central Florida, 
  Orlando, FL 32816, USA\\
  {\tt \{ericaryo,loganlebanoff\}@knights.ucf.edu\,\, feiliu@cs.ucf.edu}}

\date{}

\begin{document}
\maketitle
\begin{abstract}

Generating an abstract from a collection of documents is a desirable capability for many real-world applications. 
However, abstractive approaches to multi-document summarization have not been thoroughly investigated.
This paper studies the feasibility of using Abstract Meaning Representation (AMR), a semantic representation of natural language grounded in linguistic theory, as a form of content representation.
Our approach condenses source documents to a set of summary graphs following the AMR formalism.
The summary graphs are then transformed to a set of summary sentences in a surface realization step.  
The framework is fully data-driven and flexible.
Each component can be optimized independently using small-scale, in-domain training data.
We perform experiments on benchmark summarization datasets and report promising results. 
We also describe opportunities and challenges for advancing this line of research.

\end{abstract}

\section{Introduction}
\label{sec:intro}

Abstractive summarization seeks to generate concise and grammatical summaries that preserve the meaning of the original; further, they shall abstract away from the source syntactic forms.
The task often involves high-level text transformations such as sentence fusion, generalization, and paraphrasing~\cite{Jing:1999}.
Recent neural abstractive summarization studies focus primarily on single-document summarization~\cite{Paulus:2017,See:2017}. 
These approaches are limited by the availability of training data, and large datasets for multi-document summarization can be costly to obtain.
Generating abstractive summaries for sets of source documents thus remains a challenging task.

Traditional approaches to abstractive summarization often condense the source documents to a set of ``semantic units,'' then reconstruct abstractive summaries from these semantic units.
Previous work has investigated various forms of content representation. 
Examples include noun/verb phrases~\cite{Genest:2011,Bing:2015}, word-occurrence graphs~\cite{Ganesan:2010}, syntactic parse trees~\cite{Cheung:2014,Gerani:2014}, and domain-specific templates~\cite{Pighin:2014}.
Nonetheless, generating summary text from these heuristic forms of representation can be difficult.
There is an increasing need to exploit a semantic formalism so that condensing source documents to this form and generating summary sentences from it can both be carried out in a principled way.

\begin{figure}[t]
\begin{center}
\includegraphics [width=5.5in] {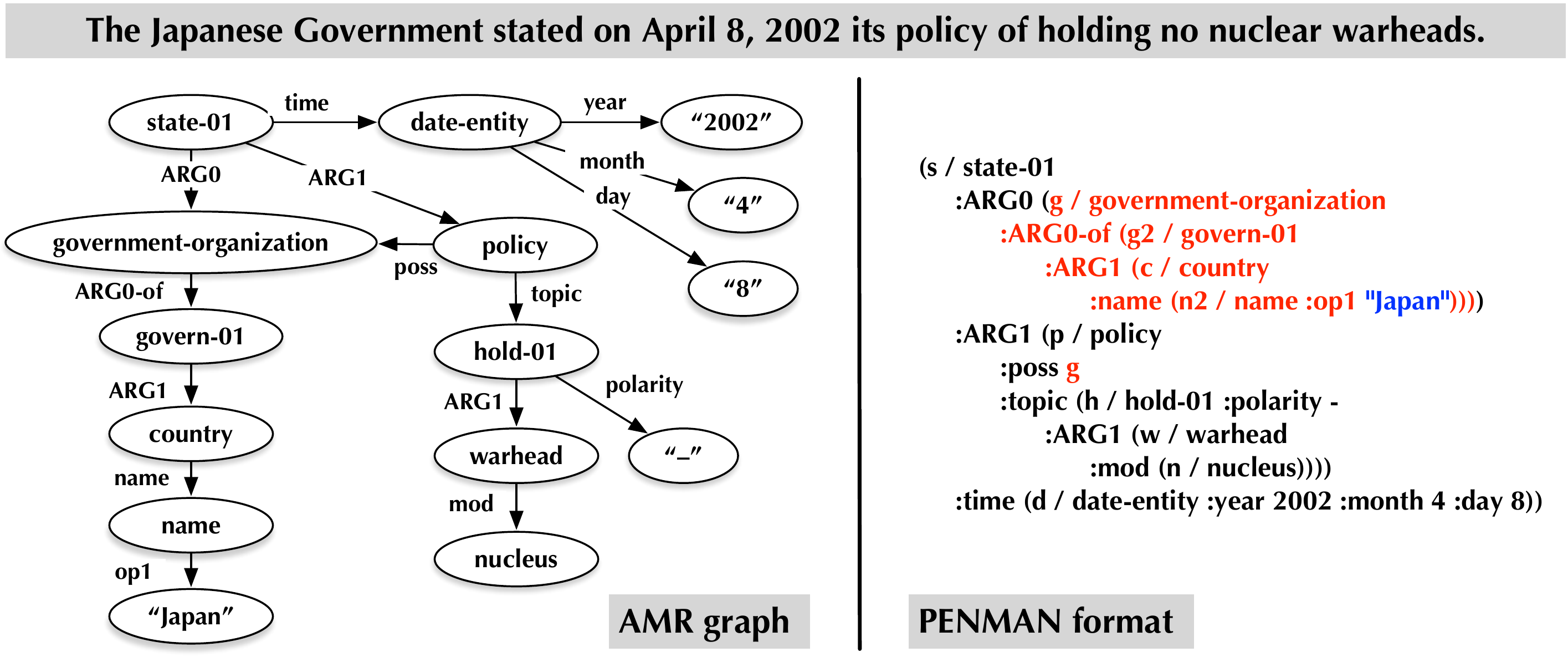}
\caption{A example sentence, its goldstandard AMR graph, and the corresponding PENMAN format.}
\label{fig:amr_graph}
\end{center}
\vspace{-0.2in}
\end{figure}

This paper explores Abstract Meaning Representation (AMR, Banarescu et al., 2013\nocite{Banarescu:2013}) as a form of content representation.
AMR is a semantic formalism based on propositional logic and the neo-Davidsonian event representation~\cite{Parsons:1990,Schein:1993}.
It represents the meaning of a sentence using a rooted, directed, and acyclic graph, where nodes are concepts and edges are semantic relations.
Figure~\ref{fig:amr_graph} shows an example AMR graph.
A concept node can be a PropBank frameset (``state-01''), an English word (``warhead''), a special keyword (``date-entity''), or a string literal (``Japan'').
A relation can be either a core argument (``ARG0,'' ``ARG1'') or a modification relationship (``mod,'' ``time'' ). 
The AMR representation abstracts away from surface word strings and syntactic structure, producing a language-neutral representation of meaning.
The graph representation is flexible and not specifically designed for a particular domain. 
It is thus conceptually appealing to explore AMR for abstractive summarization.

Our goal in this work is to generate a text abstract containing multiple sentences from a cluster of news articles discussing a given topic.
The system framework includes three major components:
\textit{source sentence selection} takes a set of news articles as input and selects sets of similar sentences covering different aspects of the topic;
\textit{content planning} consumes a set of similar sentences and derives a summary graph from them;
\textit{surface realization} transforms a summary graph to a natural language sentence.
This framework allows each component (source sentence selection, content planning, surface realization) to be individually optimized using small-scale, in-domain training data, reducing the need for large-scale parallel training data.
Our research contributions are summarized as follows:
\begin{itemize}[topsep=3pt,itemsep=-1pt,leftmargin=*]

\item we investigate AMR, a linguistically-grounded semantic formalism, as a new form of content representation for multi-document summarization.
Liu et al.~\shortcite{Liu:2015:NAACL} conducted a pilot study using AMR for single-document summarization. This paper exploits the structured prediction framework but presents a full pipeline for generating abstractive summaries from multiple source documents;

\item we study to what extent the AMR parser and generator, used for mapping text to and from AMR, can impact the summarization performance. We also compare multiple source sentence selection strategies to group source sentences into clusters covering various aspects of the topic;

\item we conduct extensive experiments on benchmark summarization datasets, and contrast our work with state-of-the-art baselines, including the pointer-generator networks~\cite{See:2017}.
Results show that leveraging the AMR representation for summarization is promising. Our framework is flexible, allowing different components to be optimized independently using small-scale, in-domain datasets.
We finally describe opportunities and challenges for advancing this line of research.

\end{itemize}

\section{Related Work}
\label{sec:related_work}

Neural abstractive summarization has sparked great interest in recent years.
These approaches focus primarily on short text summarization and single-document summarization~\cite{Rush:2015,Nallapati:2016}.
Variants of the neural encoder-decoder architecture have been exploited to reduce word repetitions~\cite{See:2017,Suzuki:2017}, improve the attention mechanism~\cite{Chen:2016,Zhou:2017,Tan:2017}, control the summary length~\cite{Kikuchi:2016}, reduce the occurrence of out-of-vocabulary tokens in summaries~\cite{See:2017}, improve the learning objective and search~\cite{Ranzato:2016,Huang:2017}, and generate summaries that are true to the original inputs~\cite{Cao:2018,Song:2018}.
Training neural models generally requires large amounts of data; they are often acquired by pairing news articles with titles or human-written highlights.
Nonetheless, obtaining parallel data for multi-document summarization is often costly.
There is thus a need to investigate alternative approaches that are less data-thirsty.

Abstractive summarization via natural language generation (NLG, Reiter and Dale, 2000; Gatt and Krahmer, 2018\nocite{Reiter:2000,Gatt:2018}) is a promising line of work.
The approaches often identify salient text units from source documents, arrange them in a compact form, such as domain-specific templates, and subsequently synthesize them into natural language texts~\cite{Barzilay:1999,Genest:2011,Oya:2014:INLG,Gerani:2014,Fabbrizio:2014}.
A challenge faced by these approaches is that there lacks a principled means of content representation. 
This paper studies the feasibility of using AMR, a semantic formalism grounded in linguistic theory, for content representation.
Within this framework, condensing source documents to summary AMR graphs and generating natural language sentences from summary graphs are both data-driven and not specifically designed for any domain.

The AMR formalism has demonstrated great potential on a number of downstream applications, including machine translation~\cite{Tamchyna:2015:S2MT}, entity linking~\cite{Pan:2015:NAACL}, summarization~\cite{Liu:2015:NAACL,Takase:2016}, question answering~\cite{Jurczyk:2015}, and machine comprehension~\cite{Sachan:2016}.
Moreover, significant research efforts are dedicated to map English sentences to AMR graphs~\cite{Flanigan:2014,Wang:2015:ACL,Wang:2015:NAACL,Ballesteros:2017,Buys:2017,Damonte:2017,Szubert:2018}, and generating sentences from AMR~\cite{Flanigan:2016,Song:2016,Pourdamghani:2016,Song:2017:ACL,Konstas:2017}.
These studies pave the way for further research exploiting AMR for multi-document summarization.

\section{Our Approach}
\label{sec:framework}

\begin{figure*}
\begin{center}
\includegraphics [width=6.2in] {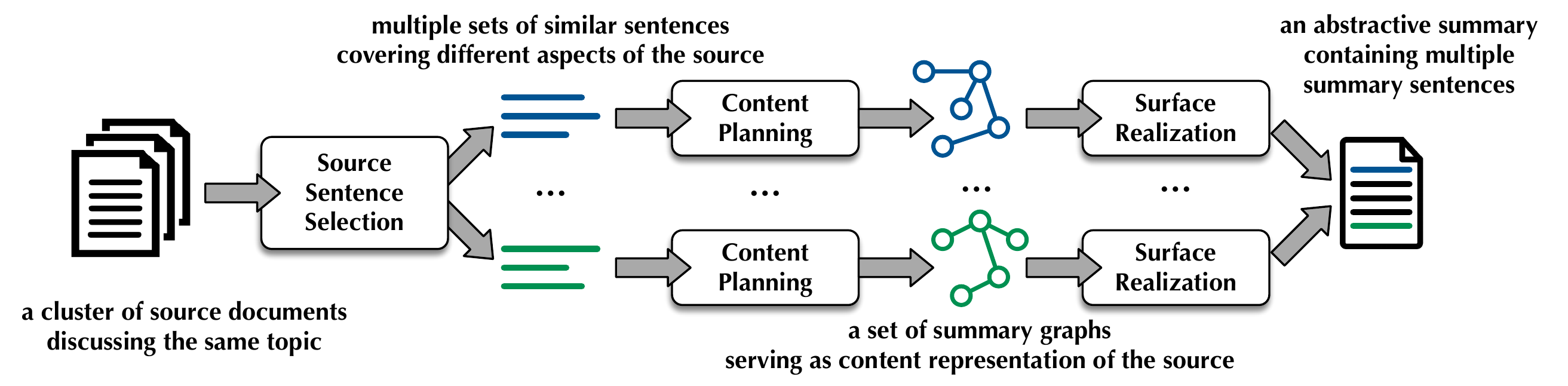}
\caption{Our system framework, consisting of three major components.}
\label{fig:framework}
\end{center}
\vspace{-0.2in}
\end{figure*}

We describe our major system components in this section.
In particular, content planning (\S\ref{sec:content_planning}) takes as input a set of similar sentences.
It maps each sentence to an AMR graph, merges all AMR graphs to a connected \textit{source graph}, then extracts a \textit{summary graph} from the source graph via structured prediction.
Surface realization (\S\ref{sec:surface_realization}) converts a summary graph to its PENMAN representation~\cite{Banarescu:2013} and generates a natural language sentence from it.
Source sentence extraction (\S\ref{sec:sentence_selection}) selects sets of similar sentences from source documents discussing different aspects of the topic.
The three components form a pipeline to generate an abstractive summary from a collection of documents.
Figure~\ref{fig:framework} illustrates the system framework.
In the following sections we describe details of the components.

\subsection{Content Planning}
\label{sec:content_planning}

The meaning of a source sentence is represented by a rooted, directed, and acyclic AMR graph~\cite{Banarescu:2013}, where nodes are concepts and edges are semantic relations.
A sentence AMR graph can be obtained by applying an AMR parser to a natural language sentence.
In this work we investigate two AMR parsers to understand to what extent the performance of AMR parsing may impact the summarization task. 
JAMR~\cite{Flanigan:2014} presents the first open-source AMR parser.
It introduces a two-part algorithm that first identifies concepts from the sentence and then determines the relations between them by searching for the maximum spanning connected subgraph (\textsc{mscg}) from a complete graph representing all possible relations between the identified concepts. 
CAMR~\cite{Wang:2015:NAACL} approaches AMR parsing from a different perspective. 
It describes a transition-based AMR parsing algorithm that transforms from a dependency parse tree to an AMR graph.
We choose JAMR and CAMR because these parsers have been made open-source and both of them reported encouraging results in the recent SemEval evaluations~\cite{May:2016,May:2017}.

\vspace{0.05in}
\noindent \textbf{Source Graph Construction.} 
Given a set of source sentences and their AMR graphs, source graph construction attempts to consolidate all sentence AMR graphs to a connected \textit{source graph}.
This is accomplished by performing \textit{concept merging}.
Graph nodes representing the same concept, determined by the surface word form, are merged to a single node in the source graph. 
Importantly, we perform coreference resolution on the source documents to identify clusters of mentions of the same entity or event. 
Graph nodes representing these mentions are also merged.
A special treatment to date entity (see ``date-entity'' in Figure~\ref{fig:amr_graph}) and named entity (``country'') includes collapsing the subtree to a ``mega-node'' whose surface form is the concatenation of the consisting concepts and relations (e.g., ``date-entity\_:year\_2002\_:month\_1\_:day\_5''). 
These mega-nodes can then only be merged with other identical fragments.
Finally, a `ROOT' node is introduced; it is connected to the root of each sentence AMR graph, yielding a connected source graph.

\vspace{0.05in}
\noindent \textbf{Summary Graph Extraction.} 
We hypothesize that a summary graph, containing the salient content of source texts, can be identified from the source graph via a trainable, feature-rich structured prediction framework.
The framework iteratively performs \textit{graph decoding} and \textit{parameter update}. 
The former identifies an optimal summary graph using integer linear programming, while the latter performs parameter update by minimizing a loss function that measures the difference between the system-decoded summary graph and the goldstandard summary graph.

We use $G = (V, E)$ to represent the source graph. 
Let $v_i$ and $e_{i,j}$ be a set of binary variables where $v_i = 1$ (or $e_{i,j} = 1$) indicates the corresponding source graph node (or edge) is selected to be included in the summary graph.
The node and edge saliency are characterized by a set of features, represented using $\mathbf{f}(i)$ and $\mathbf{g}(i,j)$, respectively.
$\boldsymbol\theta$ and $\boldsymbol\phi$ are the feature weights.
Eq.~(\ref{equ:scoring_func}) presents a scoring function for any graph $G$. 
It can be factorized into a sum of scores for selected nodes and edges.
In particular, $[\boldsymbol\theta^\top \mathbf{f}(i)]_{v_i=1}$ denotes the node score (if $v_i$ is selected) and $[\boldsymbol\phi^\top \mathbf{g}(i,j)]_{e_{i,j}=1}$ denotes the edge score.
\begin{align*}
\label{equ:scoring_func}
\displaystyle
\mathit{score}(G; \boldsymbol\theta, \boldsymbol\phi) 
= \sum_{i=1}^N v_i \underbrace{[\boldsymbol\theta ^\top \mathbf{f}(i)]_{v_i}}_{\mbox{node score}} + \sum_{(i,j) \in E} e_{i,j} \underbrace{[\boldsymbol\phi^\top \mathbf{g}(i,j)]_{e_{i,j}}}_{\mbox{edge score}}
\numberthis
\end{align*}

Features characterizing the graph nodes and edges are adopted from~\cite{Liu:2015:NAACL}.
They include concept/relation labels and their frequencies in the documents, average depth of the concept in sentence AMR graphs, position of sentences containing the concept/relation, whether the concept is a named entity/date entity, and the average length of concept word spans.
We additionally include the concept TF-IDF score and if the concept occurs in a major news event~\cite{Wiki:2018:Events}.
All features are binarized.

The \textit{graph decoding} process searches for the summary graph that maximizes the scoring function: $\bm\hat{G} = \argmax_G \mathit{score}(G)$.
We can formulate graph decoding as an integer linear programming problem. 
Each summary graph corresponds to a set of values assigned to the binary variables $v_i$ and $e_{i,j}$. 
We implement a set of linear constraints to ensure the decoded graph is connected, forms a tree structure, and limits to $L$ graph nodes~\cite{Liu:2015:NAACL}.

The \textit{parameter update} process adjusts $\boldsymbol\theta$ and $\boldsymbol\phi$ to minimize a loss function capturing the difference between the system-decoded summary graph ($\bm\hat{G}$) and the goldstandard summary graph ($G^\ast$).
Eq.~(\ref{equ:loss_func}) presents the structured perceptron loss. 
Minimizing this loss function with respect to $\boldsymbol\theta$ and $\boldsymbol\phi$ is straightforward.
However, the structured perceptron loss has problems when the goldstandard summary graph $G^\ast$ is unreachable via the graph decoding process.
In that case, there remains a gap between  $\mathit{score}(\bm\hat{G})$ and $\mathit{score}(G^\ast)$ and the loss cannot be further minimized.
The structured ramp loss (Eq.~(\ref{equ:ramp_loss})) addresses this problem by performing cost-augmented decoding. 
It introduces a cost function $\mathit{cost}(G;G^\ast)$ that can also be factored over graph nodes and edges.
A cost of 1 is incurred if the system graph and the goldstandard graph disagree on whether a node (or edge) should be included.
As a result, the first component of the loss function $\max_{G} (\mathit{score}(G) + \mathit{cost}(G; G^\ast))$ yields a decoded system graph that is slightly \textit{worse} than $\bm\hat{G}=\argmax_{G} \mathit{score}(G)$;
and the second component $\max_{G} (\mathit{score}(G) - \mathit{cost}(G; G^\ast))$ yields a graph that is slightly \textit{better} than $\bm\hat{G}$.
The scoring difference between the two system graphs becomes the structured ramp loss we wish to minimize.
This formulation is similar to the objective of the structured support vector machines (SSVMs, Tsochantaridis et al., 2005\nocite{Tsochantaridis:2005}).
Becase decent results have been reported by~\cite{Liu:2015:NAACL}, we adopt structured ramp loss in all experiments.
\begin{align*}
\displaystyle
\mathcal{L}_{\mbox{\small{perc}}}(\boldsymbol\theta, \boldsymbol\phi) 
&= \max_G \mathit{score}(G) - \mathit{score}(G^\ast)
= \mathit{score}(\bm\hat{G}) - \mathit{score}(G^\ast)
\numberthis\label{equ:loss_func}\\
\mathcal{L}_{\mbox{\small{ramp}}}(\boldsymbol\theta, \boldsymbol\phi) &= \max_{G} (\mathit{score}(G) + \mathit{cost}(G; G^\ast))
- \max_{G} (\mathit{score}(G) - \mathit{cost}(G; G^\ast))
\numberthis\label{equ:ramp_loss}
\end{align*}

\subsection{Surface Realization}
\label{sec:surface_realization}

\setlength{\intextsep}{1pt}%

\begin{wrapfigure}{r}{0.5\textwidth}
\centering
\begin{minipage}[b]{0.5\textwidth}
\begin{algorithm}[H]
\begin{algorithmic}[1]
\REQUIRE Triples: (src\_concept, relation, tgt\_concept).
\STATE $r \leftarrow$ index of the ROOT concept
\STATE $\mathcal{P} \leftarrow$ all paths from ROOT to leaves, sorted by the concept indices on the paths
\STATE \COMMENT{\textcolor{darkblue}{Set all concepts and relations as unvisited}}
\STATE $\mathit{visited}[\mathcal{C}_m]$ $\leftarrow$ \textsc{false}, $\forall m$
\STATE $\mathit{visited}[\mathcal{R}_{m,n}]$ $\leftarrow$ \textsc{false}, $\forall m, n$
\FOR{$i=1,\ldots,|\mathcal{P}|$}
\STATE $\mathit{flag\_special\_concept} \leftarrow$ \textsc{false}.
\STATE $k \leftarrow 0$
\WHILE{$k < |p_i|$}
\STATE $k \leftarrow k+1$
\STATE $n' \leftarrow$ index of the $k$-th concept on the path
\STATE $m' \leftarrow$ index of the previous concept
\IF{$\mathit{visited}[\mathcal{C}_{n'}]$ = \textsc{false}}
\STATE \COMMENT{\textcolor{darkblue}{Concept unvisited}}
\STATE $\mathit{visited}[\mathcal{C}_{n'}]$ $\leftarrow$ \textsc{true}
\STATE $\mathit{visited}[\mathcal{R}_{m',n'}]$ $\leftarrow$ \textsc{true}
\STATE $\mathit{output}$ += $(k-1)\ast$TAB
\IF{$\mathcal{C}_{n'}$ is a string literal}
\STATE $\mathit{output}$ += $\mathcal{R}_{m',n'}$::$\mathcal{C}_{n'}$
\STATE $\mathit{flag\_special\_concept} \leftarrow$ \textsc{true}
\STATE \textsc{break}
\ELSIF{$k < |p_i|$}
\STATE $\mathit{output}$ += $\mathcal{R}_{m',n'}$::``(''::$\mathcal{C}_{n'}$::EOS
\ELSE 
\STATE $\mathit{output}$ += $\mathcal{R}_{m',n'}$::``(''::$\mathcal{C}_{n'}$
\ENDIF
\ELSIF{$\mathit{visited}[\mathcal{R}_{m',n'}]$ = \textsc{false}}
\STATE \COMMENT{\textcolor{darkblue}{Concept reentrance}}
\STATE $\mathit{visited}[\mathcal{R}_{m',n'}] \leftarrow$ \textsc{true}
\STATE $\mathit{output}$ += $(k-1)\ast$TAB
\STATE $\mathit{output}$ += $\mathcal{R}_{m',n'}$::$\mathcal{C}_{n'}$
\STATE $\mathit{flag\_special\_concept} \leftarrow$ \textsc{true}
\STATE \textsc{break}
\ENDIF
\ENDWHILE
\STATE \COMMENT{\textcolor{darkblue}{Output path ending brackets and EOS.}}
\STATE $k' \leftarrow$ tracing the path backwards to find position of the closest ancestor who has an unvisited child; if none exists, $k' \leftarrow 0$
\IF{$\mathit{flag\_special\_concept}$ = \textsc{true}}
\STATE $\mathit{output}$ += $(k-k'-1)\ast$``)''::EOS
\ELSE
\STATE $\mathit{output}$ += $(k-k')\ast$``)''::EOS
\ENDIF
\ENDFOR
\end{algorithmic}
\caption{An algorithm for transforming a summary graph to the PENMAN format.}
\label{alg:graph_to_penman}
\end{algorithm}
\end{minipage}
\vspace{-0.2in}
\end{wrapfigure}

The surface realization component converts each summary graph to a natural language sentence. 
This is a nontrivial task, because AMR abstracts away from the source syntactic forms and an AMR graph may correspond to a number of valid sentence realizations.
In this section we perform two subtasks that have not been investigated in previous studies.
We first convert the summary AMR graphs to the PENMAN format (Figure~\ref{fig:amr_graph}), which is a representation that can be understood by humans. It is also the required input form for an AMR-to-text generator.
We then leverage the AMR-to-text generator to generate English sentences from summary graphs.

Algorithm~\ref{alg:graph_to_penman} presents our algorithm for transforming a summary graph to the PENMAN format.
Because the summary graph is rooted and acyclic, we can extract all paths from the ROOT node to all leaves.
These paths are sorted by the concept indices on the path, and the paths sharing the same ancestors will be processed in order. 
The core of Algorithm~\ref{alg:graph_to_penman} is the \textit{while} loop (line 9--35); it writes out one concept per line.
An AMR concept can have three forms: a regular form (``c / country''), string literal (``Japan''), or a re-entrance (``g'').
The last two are treated as ``special forms.'' 
In these cases, a relation is first written out, followed by the special concept ($\mathcal{R}_{m',n'}$::$\mathcal{C}_{n'}$). ``::'' denotes a whitespace.
A regular concept will be wrapped by a left bracket ($\mathcal{R}_{m',n'}$::``(''::$\mathcal{C}_{n'}$, line 23/25) and a right bracket ($(k-k')\ast$``)''::EOS, line 41).
Finally, a proper number of closing brackets is postpended to each path (line 37--42).

To transform the PENMAN string to a summary sentence we employ the JAMR AMR-to-text generator~\cite{Flanigan:2016}. 
JAMR is the first full-fledged AMR-to-text generator.
It is trained on approximately 10K sentences and achieves about 22\% BLEU score on the test set.
The system first transforms the input graph to a spanning tree, and then decodes it into a string using a tree-to-string transducer and a language model.
The final output sentence is the highest-scoring sentence according to a feature-rich discriminatively trained linear model.
We choose the JAMR AMR-to-text generator because of its competitive performance in the recent SemiEval evaluations~\cite{May:2017}.

\subsection{Source Sentence Selection}
\label{sec:sentence_selection}

We seek to generate an abstractive summary containing multiple sentences from a cluster of documents discussing a single topic (e.g., health and safety). 
Each summary sentence will cover a topic aspect; it is generated by fusing a set of relevant source sentences.
We thus perform clustering on all source sentences to find salient topic aspects and their corresponding sets of similar sentences. 
Spectral clustering has been shown to perform strongly on different clustering problems~\cite{Ng:2002,Yogatama:2009}.
The approach constructs an affinity matrix by applying a pairwise similarity function to all source sentences.
It then calculates the eigenvalues of the matrix and performs clustering in the low-dimensional space spanned by the largest eigenvectors.
A large cluster indicates a salient topic aspect.
We focus on the $M$ largest clusters and extract $N$ sentences from each cluster.\footnote{We use $N$=$M$=5 in our experiments. This setting fuses 5 source sentences to a summary sentence. 
It then produces 5 summary sentences for each topic, corresponding to the average number of sentences in human summaries.}
These sentences have the highest similarity scores with other sentences in the cluster.
The selected sets of relevant sentences are later fed to the content planning component to generate summary AMR graphs.

Training the content planning component, however, requires sets of source sentences paired with their goldstandard summary graphs.
Manually selecting sets of sentences and annotating summary graphs is costly and time-consuming. 
Instead, we leverage human reference summaries to create training instances.
We obtain summary graphs by AMR-parsing sentences of human reference summaries.
For every reference sentence, we further extract a set of source sentences. They are judged similar to the reference sentence via a similarity metric.
The summary AMR graphs and sets of source sentences thus form the training data for content planning.
We gauge how best to select source sentences by exploring different similarity metrics.
In particular,
(i) \textbf{LCS} calculates the longest common subsequence between a candidate source sentence and the reference sentence;
(ii) \textbf{VSM} represents sentences using the vector space model and calculates the cosine similarity between the two sentence vectors; 
(iii) \textbf{Smatch}~\cite{Cai:2013} calculates the F-score of AMR concepts between the candidate and reference sentences;
(iv) \textbf{Concept Coverage} selects source sentences to maximize the coverage of AMR concepts of the reference sentence.
We experiment with these source sentence selection strategies and compare their effectiveness in Section \S\ref{sec:results_graph_prediction}.

\section{Datasets and Baselines}
\label{sec:datasets}

We perform experiments on standard multi-document summarization datasets\footnote{\url{https://duc.nist.gov/data.html} \quad \url{https://tac.nist.gov/data/index.html} }, prepared by the NIST researchers for DUC/TAC competitions and later exploited by various summarization studies~\cite{Nenkova:2011,Hong:2014,Yogatama:2015:EMNLP}.
A summarization instance includes generating a text summary containing 100 words or less from a cluster of 10 source documents discussing a single topic.
4 human reference summaries are provided for each cluster of documents; they are created by NIST assessors.
We use the datasets from DUC-03, DUC-04, TAC-09, TAC-10, and TAC-11 in this study, containing 30/50/44/46/44 clusters of documents respectively.

We compare our AMR summarization framework with a number of extractive (\textit{ext}-$\ast$) and abstractive (\textit{abs}-$\ast$) summarization systems, including the most recent neural encoder-decoder architecture~\cite{See:2017}.
They are described as follows.

\begin{itemize}[topsep=3pt,itemsep=-1pt]
\item \textit{ext-}\textbf{LexRank}~\cite{Erkan:2004} is a graph-based approach that computes sentence importance based on the concept of eigenvector centrality in a graph representation of source sentences;
\item \textit{ext-}\textbf{SumBasic}~\cite{Vanderwende:2007} is an extractive approach that assumes words occurring frequently in a document cluster have a higher chance of being included in the summary;
\item \textit{ext-}\textbf{KL-Sum}~\cite{Haghighi:2009} describes a method that greedily adds sentences to the summary so long as it decreases the KL divergence;
\item \textit{abs-}\textbf{Opinosis}~\cite{Ganesan:2010} generates abstractive summaries by searching for salient paths on a word co-occurrence graph created from source documents;
\item \textit{abs-}\textbf{Pointer-Generator}~\cite{See:2017} describes a neural encoder-decoder architecture. 
It encourages the system to copy words from the source text via pointing, while retaining the ability to produce novel words through the generator. 
It also includes a coverage mechanism to keep track of what has been summarized, thus reducing word repetition.
The pointer-generator networks have not been tested for multi-document summarization. In this study we evaluate their performance on the DUC/TAC datasets.
\end{itemize}

\begin{table}[t]
\setlength{\tabcolsep}{5pt}
\centering
\begin{tabular}{|l|ccc|ccc|ccc|ccc|}
\hline
& \multicolumn{3}{c|}{\textbf{Nodes}} & \multicolumn{3}{c|}{\textbf{(Oracle) Nodes}} & \multicolumn{3}{c|}{\textbf{Edges}} & \multicolumn{3}{c|}{\textbf{(Oracle) Edges}}\\
\textbf{Approach} & \textbf{P} & \textbf{R} & \textbf{F} & \textbf{P} & \textbf{R} & \textbf{F} & \textbf{P} & \textbf{R} & \textbf{F} & \textbf{P} & \textbf{R} & \textbf{F}\\
\hline
\hline
LCS & 16.7 & 26.7 & 19.9 & 31.5 & 49.5 & 37.6 & 6.7 & 8.0 & 6.9 & 16.1 & 18.7 & 16.6\\
Smatch & 20.9 & 33.2 & 24.9 & 33.2 & 52.0 & 39.6 & 9.3 & 10.7 & 9.4 & 17.2 & 20.1 & 17.8\\
Concept Cov. & \textbf{25.0} & \textbf{40.3} & \textbf{30.1} & \textbf{48.8} & \textbf{77.5} & \textbf{58.7} & 7.3 & 10.0 & 8.0 & 18.9 & \textbf{25.3} & 20.8\\
VSM & 24.0 & 38.6 & 28.8 & 40.8 & 64.3 & 48.9 & \textbf{9.6} & \textbf{11.3} & \textbf{9.8} & \textbf{21.1} & 25.1 & \textbf{22.1}\\
\hline
\end{tabular}
\caption{
Summary graph prediction results on the DUC-04 dataset. 
The scores measure how well the predicted summary graphs match reference summary graphs on nodes and edges.
Reference summary graphs are created by parsing reference summary sentences using the CAMR parser. 
``Oracle'' results are obtained by performing only cost-based decoding. 
They establish an upper bound for the respective approaches.
}
\label{tab:results_graph_nodes}
\vspace{-0.2in}
\end{table}

\section{Experimental Results}
\label{sec:experiments}

In this section we evaluate our AMR summarization framework.
We are interested in knowing how well the system performs on predicting summary graphs from sets of relevant source sentences (\S\ref{sec:results_graph_prediction}). 
We also investigate the system's performance on generating abstractive summaries and its comparison with various baselines (\S\ref{sec:results_summarization}).
Finally, we provide an analysis on system summaries and outline challenges and opportunities for advancing this line of work (\S\ref{sec:results_analysis}).

\subsection{Results on Summary Graph Prediction}
\label{sec:results_graph_prediction}

Graph prediction results on the DUC-04 dataset (trained on DUC-03) are presented in Table~\ref{tab:results_graph_nodes}.
We report how well the decoded summary graphs match goldstandard summary graphs on  nodes (concepts) and edges (relations).
We compare several strategies to select sets of source sentences.
The goldstandard summary graphs are created by parsing the reference summary sentences via the CAMR parser~\cite{Wang:2015:NAACL}.
Note that we cannot obtain goldstandard summary graphs for sets of source sentences selected by spectral clustering.
This approach therefore is not evaluated for graph prediction.
The system-decoded summary graphs are limited to 15 graph nodes, corresponding to the average number of words in reference summary sentences (stopwords excluded). 
We additionally report ``Oracle'' decoding results, obtained by performing only cost-based decoding $\textstyle \bm\hat{G} = \argmax_{G} (- \mathit{cost}(G; G^\ast))$ on the source graph, where $G^\ast$ is the goldstandard summary graph. 
The oracle results establish an upper bound for the respective approaches.

We observe that node prediction generates better results than edge prediction.
Using `Concept Cov,' the system-decoded summary graphs successfully preserve 40.3\% of the goldstandard summary concepts, and this number increases to 77.5\% when using oracle decoding, indicating the content planning component is effective at identifying important source concepts and preserving them in summary graphs.
`VSM' performs best on edge prediction.
It achieves an F-score of 9.8\% and the oracle decoding further boosts the performance to 22.1\%.
We observe that only 42\% of goldstandard summary bigrams appear in the source documents, serving as a cap for edge prediction.
The results suggest that `VSM' is effective at selecting sets of source sentences containing salient source relations.
The high performance on summary node prediction but low on edge prediction suggests that future work may consider increasing the source graph connectivity by introducing edges between concepts so that salient summary edges can be effectively preserved.

\begin{table*}[t]
\setlength{\tabcolsep}{5pt}
\centering
\begin{tabular}{|l|l|rrr|rrr|rrr|}
\hline
\multicolumn{2}{|c|}{} & \multicolumn{3}{c|}{\textbf{ROUGE-1}} & \multicolumn{3}{c|}{\textbf{ROUGE-2}} & \multicolumn{3}{c|}{\textbf{ROUGE-SU4}}\\
& \textbf{System} & \multicolumn{1}{c}{\textbf{P}} & \multicolumn{1}{c}{\textbf{R}} & \multicolumn{1}{c|}{\textbf{F}} & \multicolumn{1}{c}{\textbf{P}} & \multicolumn{1}{c}{\textbf{R}} & \multicolumn{1}{c|}{\textbf{F}} & \multicolumn{1}{c}{\textbf{P}} & \multicolumn{1}{c}{\textbf{R}} & \multicolumn{1}{c|}{\textbf{F}}\\
\hline
\hline
\multirow{7}{*}{\textbf{DUC 2004}} 

& \textit{ext}-SumBasic & 37.5 & 24.9 & 29.5 & 5.3 & 3.6 & 4.3 & 11.1 & 7.3 & 8.6\\
& \textit{ext}-KL-Sum & 31.1 & 31.1 & 31.0 & 6.0 & 6.1 & 6.0 & 10.2 & 10.3 & 10.2\\
& \textit{ext}-LexRank & 34.3 & 34.6 & 34.4 & 7.1 & 7.2 & 7.1 & 11.1 & 11.2 & 11.2\\
\cline{2-11}
& \textit{abs}-Opinosis & {36.5} & 23.7 & 27.5 & 7.2 & 4.3 & 5.1 & {11.7} & 7.4 & 8.6\\
& \textit{abs}-Pointer-Gen-all & 37.5 & 20.9 & 26.5 & 8.0 & 4.4 & \textbf{5.6} & 12.3 & 6.7 & 8.5\\
& \textit{abs}-Pointer-Gen & 33.2 & 21.5 & 25.6 & 5.8 & 3.8 & 4.5 & 10.3 & 6.6 & 7.9\\
& \textit{abs}-AMRSumm-Clst & 29.9 & 30.5 & \textbf{30.2} & 4.1 & 4.2 & 4.1 & 8.7 & 8.9 & \textbf{8.8}\\
& \textit{abs}-AMRSumm-VSM & 36.7 & {39.0} & \textbf{37.8} & 6.5 & 6.9 & \textbf{6.6} & 11.4 & {12.2} & \textbf{11.8}\\
\hline
\hline
\multirow{7}{*}{\textbf{TAC 2011}} 
& \textit{ext}-SumBasic & 37.3 & 28.2 & 31.6 & 6.9 & 5.5 & 6.1 & 11.8 & 9.0 & 10.1\\
& \textit{ext}-KL-Sum  & 31.2 & 31.4 & 31.2 & 7.1 & 7.1 & 7.1 & 10.5 & 10.6 & 10.6\\
& \textit{ext}-LexRank & 32.9 & 33.3 & 33.1 & 7.4 & 7.6 & 7.5 & 11.1 & 11.2 & 11.1\\
\cline{2-11}
& \textit{abs}-Opinosis & 38.0 & 20.4 & 25.2 & {8.6} & 4.0 & 5.1 & 12.9 & 6.5 & 8.1\\
& \textit{abs}-Pointer-Gen-all & 37.3 & 22.2 & 27.6 & 7.8 & 4.6 & \textbf{5.8} & 12.2 & 7.1 & 8.9\\
& \textit{abs}-Pointer-Gen & 34.4 & 21.6 & 26.2 & 6.9 & 4.4 & 5.3 & 10.9 & 6.8 & 8.2\\
& \textit{abs}-AMRSumm-Clst & 32.2 & 31.7 & \textbf{31.9} & 4.7 & 4.7 & 4.7 & 9.8 & 9.7 & \textbf{9.7}\\
& \textit{abs}-AMRSumm-VSM & {40.1} & {42.3} & \textbf{41.1} & 8.1 & \textbf{8.5} & \textbf{8.3} & {13.1} & {13.9} & \textbf{13.5}\\
\hline
\end{tabular}
\caption{
Summarization results on DUC-04 and TAC-11 datasets. 
We compare the AMR summarization framework (AMRSumm-*) with both extractive (\textit{ext-}*) and abstractive (\textit{abs-}*) summarization systems.
}
\label{tab:results_summarization}
\vspace{-0.1in}
\end{table*}

\subsection{Results on Summarization}
\label{sec:results_summarization}

In Table~\ref{tab:results_summarization} we report the summarization results evaluated by ROUGE~\cite{Lin:2004}.
In particular, R-1, R-2, and R-SU4 respectively measure the overlap of unigrams, bigrams, and skip bigrams (up to 4 words) between system and reference summaries.
Our AMR summarization framework outperforms all other abstractive systems with respect to R-1 and R-SU4 F-scores on both DUC-04 (trained on DUC-03) and TAC-11 (trained on TAC-09,10) datasets.
``AMRSumm-VSM'' further produces the highest R-2 F-scores.
We conjecture that the R-2 scores of AMRSumm-* are related to the performance of the AMR-to-text generator~\cite{Flanigan:2016}.
When it transforms a summary graph to text, there can be multiple acceptable realizations (e.g., ``I hit my hand on the table'' or ``My hand hit the table'') and the one chosen by the AMR generator may not always be the same as the source text.
Because abstractive systems are expected to produce novel words, they may yield slightly inferior results to the best extractive system (LexRank).
Similar findings are also reported by Nallapati et al.~\shortcite{Nallapati:2017} and See et al.~\cite{See:2017}.

We experiment with two variants of the pointer-generator networks:
``Pointer-Generator-all'' uses all source sentences of the document set and ``Pointer-Generator'' uses the source sentences selected by spectral clustering, hence the same input as ``AMRSumm-Clst.'' 
We observe that ``AMRSumm-Clst'' performs stronger than ``Pointer-Generator'' at preserving salient summary concepts, yielding R-1 F-scores of 30.2\% vs. 25.6\% and 31.9\% vs. 26.2\% on DUC-04 and TAC-11 datasets.
Further, we found that the summaries produced by the pointer-generator networks are more extractive than abstractive.
We report the percentages of summary n-grams contained in the source documents in Figure~\ref{fig:ngram}.
``Pointer-Generator-all'' has 99.6\% of unigrams, 95.2\% bigrams, and 87.2\% trigrams contained in the source documents (DUC-04).
In contrast, the ratios for human summaries are 85.2\%, 41.6\% and 17.1\%, and for ``AMRSumm-Clst'' the ratios are 84.6\%, 31.3\% and 8.4\% respectively.
Both human summaries and ``AMRSumm-Clst'' summaries tend to be more abstractive, with fewer bigrams/trigrams appeared in the source.
These results suggest that future abstractive systems for multi-document summarization may need to carefully balance between copying words from the source text with producing new words/phrases in order to generate summaries that resemble human abstracts.

\begin{table*}
\setlength{\tabcolsep}{5pt}
\centering
\begin{tabular}{|l|l|rrr|rrr|rrr|}
\hline
\multicolumn{2}{|c|}{} & \multicolumn{3}{c|}{\textbf{JAMR}} & \multicolumn{3}{c|}{\textbf{CAMR}} & \multicolumn{3}{c|}{\textbf{(Oracle) CAMR}}\\
& \textbf{Approach} & \textbf{P} & \textbf{R} & \textbf{F} & \textbf{P} & \textbf{R} & \textbf{F} & \textbf{P} & \textbf{R} & \textbf{F}\\
\hline
\hline
\multirow{3}{*}{\textbf{ROUGE-1}} & AMRSumm-Clst & 29.0 & 29.8 & 29.4 & 29.9 & 30.5 & 30.2 & 36.4 & 37.8 & 37.1\\
& AMRSumm-Concept Cov & 36.3 & 37.8 & \textbf{36.9} & 36.9 & 39.3 & \textbf{38.1} & 46.9 & 49.8 & \textbf{48.3}\\
& AMRSumm-VSM & 35.9 & 37.2 & 36.5 & 36.7 & 39.0 & 37.8 & 43.3 & 46.1 & 44.6\\
\hline
\multirow{3}{*}{\textbf{ROUGE-2}} & AMRSumm-Clst & 3.2 & 3.3 & 3.2 & 4.1 & 4.2 & 4.1 & 6.0 & 6.3 & 6.1\\
& AMRSumm-Concept Cov & 4.8 & 5.0 & \textbf{4.9} & 5.7 & 6.0 & 5.8 & 9.8 & 10.4 & \textbf{10.1}\\
& AMRSumm-VSM & 4.6 & 4.8 & 4.7 & 6.5 & 6.9 & \textbf{6.6} & 9.7 & 10.3 & 10.0\\
\hline
\end{tabular}
\caption{
Summarization results of the AMR-Summ framework on the DUC-04 dataset.
``JAMR'' uses the JAMR parser~\cite{Flanigan:2014} to produce AMR graphs for source sentences, while ``CAMR'' uses the CAMR parser~\cite{Wang:2015:NAACL}.
``Oracle'' results are obtained by performing only cost-based decoding. 
}
\label{tab:results_analysis}
\vspace{-0.1in}
\end{table*}
 
\subsection{Result Analysis}
\label{sec:results_analysis}

Table~\ref{tab:results_analysis} shows results of the AMR-Summ framework with different system configurations.
We use CAMR~\cite{Wang:2015:NAACL} to parse source sentences into AMR graphs during training, and apply either JAMR~\cite{Flanigan:2014} or CAMR to parse sentences at test time. 
We observe that the quality of AMR parsers has an impact on summarization performance.
In particular, JAMR reports 58\% F-score and CAMR reports 63\% F-score for parsing sentences.
The inter-annotator agreement places an upper bound of 83\% F-score on expected parser performance~\cite{Wang:2015:NAACL}.
There remains a significant gap between the current parser performance and the best it can achieve.
Consequently, we notice that there is a gap of 1$\sim$2 points in terms of ROUGE scores when comparing summaries produced using the two parsers. 
We notice that source sentence selection strategies making use of reference summary sentences produces better results than `AMRSumm-Clst.' Using oracle decoding further boosts the summarization performance by 7-10\% for R-1 F-score and 2-5\% for R-2 F-score. 

When examining the source and summary graphs, we notice that a simplified AMR representation could be helpful to summarization. 
As a meaning representation grounded in linguistic theory, AMR strives to be comprehensive and accurate. 
However, a summarization system may benefit from a reduced graph representation to increase the algorithm robustness.
For example, the large `semantic content units' could be collapsed to ``mega-nodes'' in some cases. 
Figure~\ref{fig:amr_analysis} shows two examples.
In the first example (``Japanese Government''), the human annotator chooses to decompose derivational morphology given that a relative clause paraphrase is possible~\cite{Schneider:2015:NAACL}. It produces 5 concept nodes, representing ``government organization that governs the country of Japan.''
In the second example, ``more than two times the amount invested in them'' also has fine-grained annotation.
For the purpose of summarization, these graph fragments could potentially be collapsed to ``mega-nodes'' and future AMR parsers may consider working on reduced AMR graphs.

\begin{table}[t]
\begin{minipage}{0.5\textwidth}
\centering
\includegraphics [width=3in] {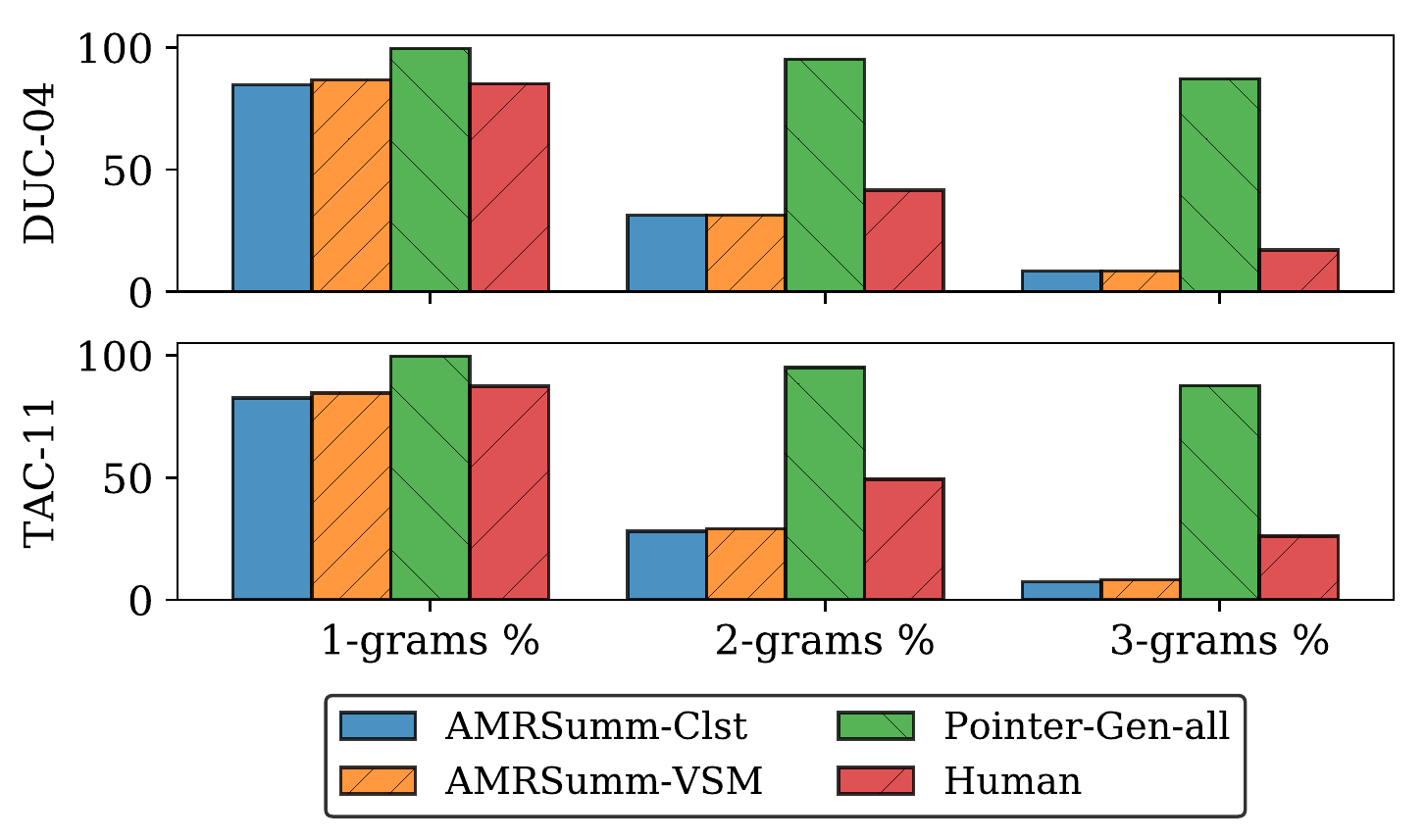}
\captionof{figure}{
Percentages of summary n-grams contained in the source documents.
Both human summaries and AMRSumm summaries are highly abstractive, with few bigrams and trigrams contained in the source.
Pointer-Generator summaries appear be more extractive than abstractive.
}
\label{fig:ngram}
\end{minipage}
\hfill
\begin{minipage}{0.48\textwidth}
\centering
\includegraphics [width=2.9in] {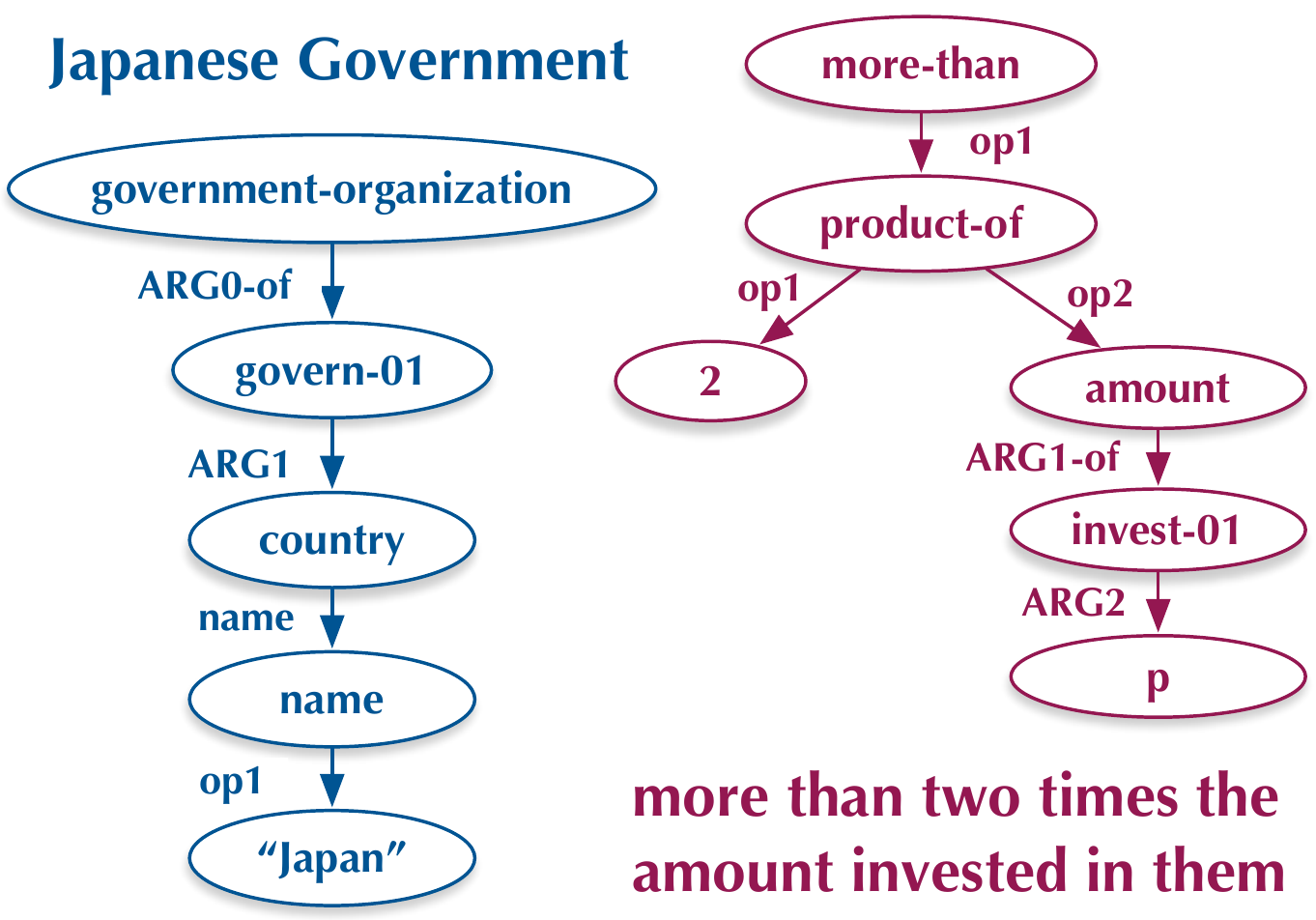}
\captionof{figure}{Example AMR for text segments.}
\label{fig:amr_analysis}
\end{minipage}
\vspace{-0.1in}
\end{table}

\section{Conclusion}
\label{sec:conclusion}

In this paper we investigated the feasibility of utilizing the AMR formalism for multi-document summarization. 
We described a full-fledged approach for generating abstractive summaries from multiple source documents.
We further conducted experiments on benchmark summarization datasets and showed that the AMR summarization framework performs competitively with state-of-the-art abstractive approaches.
Our findings suggest that the abstract meaning representation is a powerful semantic formalism that holds potential for the task of abstractive summarization.

\section*{Acknowledgements}

We are grateful to the anonymous reviewers for their insightful comments. 
The authors thank Chuan Wang, Jeffrey Flanigan, and Nathan Schneider for useful discussions.

\bibliographystyle{acl}
\bibliography{fei,amr,summ,abs_summ}

\end{document}